# Multiplier-less Artificial Neurons Exploiting Error Resiliency for Energy-Efficient Neural Computing

Syed Shakib Sarwar, Swagath Venkataramani, Anand Raghunathan, and Kaushik Roy
School of Electrical and Computer Engineering, Purdue University
{sarwar,venkata0,raghunathan,kaushik}@purdue.edu

*Abstract*— Large-scale artificial neural networks have shown significant promise in addressing a wide range of classification and recognition applications. However, their large computational requirements stretch the capabilities of computing platforms. The fundamental components of these neural networks are the neurons and its synapses. The core of a digital hardware neuron consists of multiplier, accumulator and activation function. Multipliers consume most of the processing energy in the digital neurons, and thereby in the hardware implementations of artificial neural networks. We propose an approximate multiplier that utilizes the notion of computation sharing and exploits error resilience of neural network applications to achieve improved energy consumption. We also propose Multiplier-less Artificial Neuron (MAN) for even larger improvement in energy consumption and adapt the training process to ensure minimal degradation in accuracy. We evaluated the proposed design on 5 recognition applications. The results show, 35% and 60% reduction in energy consumption, for neuron sizes of 8 bits and 12 bits, respectively, with a maximum of ~2.83% loss in network accuracy, compared to a conventional neuron implementation. We also achieve 37% and 62% reduction in area for a neuron size of 8 bits and 12 bits, respectively, under iso-speed conditions.

## I. INTRODUCTION

The human brain, which is shaped by millions of years of evolution, can easily outperform current day Von Neumann computing for a class of applications involving recognition, analytics and inference. Artificial Neural Networks (ANNs), which draw inspiration from biological neural networks, have been successfully applied to a broad spectrum of applications such as function approximation, regression analysis, pattern and sequence recognition, filtering, clustering, robotics [1-3] etc. A significant recent development in the field of ANNs is the rise of Deep Learning Networks (DLN). Using massive amounts of data and computing power, DLNs can recognize speech, interpret and classify images, read documents, and do variety of inference tasks [4-6], as well as or better than any other known algorithm. Microsoft's 'Project Adam' [7] is one such initiative that contains a DLN with more than 2 billion connections.

Due to their large computational requirements, hardware implementations of these neuromorphic architectures prove inefficient in terms of power consumption and area. Challenges of hardware implementation of ANNs have been studied from different perspectives [8,9]. One approach for pursuing efficient hardware implementation of neural networks is to modify the architecture of the networks [10-12]. To exploit the parallelism of ANNs, utilization of Graphical Processing Units (GPUs) [13, 14] has also been explored. The other approach is the use of emerging device technologies to implement neurons and synapses more efficiently. Use of hybrid memristor crossbar-array/CMOS system [15], phase-change memory devices [16, 17], resistive RAM [18], and spin based devices [19-21] in this context have been explored.

Fortunately, neural networks and their associated applications exhibit intrinsic application resilience to errors, which makes them appropriate candidates for approximate computations. Exploiting the inherent error resilience of a system, energy efficiency can be achieved by utilizing a variety of hardware [22-26] and software [27-29] techniques.

The main power hungry components of an ANN are the multipliers in the neurons which multiply inputs and corresponding synapses (weights). To address this issue, we propose an Alphabet Set Multiplier (ASM) which is approximate in nature. We utilized the Computation Sharing Multiplication (CSHM) [30-32] concept in designing the energy efficient ASM. In ASM, conventional multiplication is replaced by simplified shift and add operations. An ASM contains a pre-computer bank that generates some 'alphabets', which are lower order multiples of the input. Based on the synapse value, a proper combination of alphabet select, shift and addition operations is carried out to get the product. To achieve energy benefits, the number of 'alphabets' used in the proposed ASM are less than necessary for ideal (accurate) operation. As a result, it cannot support all the multiplication combinations. To guarantee proper functioning of the neural network, we must ensure that those unsupported multiplication combinations do not lead to significant errors during testing. For this purpose, we impose restrictions on the weights obtained from the conventionally trained network. These restrictions are similar to quantization, which drops some amount of information. As a result, accuracy loss is incurred. However, to achieve acceptable output quality, we apply retraining of the NN with restrictions in place.

The proposed ASM can replace the conventional multiplier in artificial neurons in order to get reduction in energy consumption and also gain other benefits such as reduction of area and increase in processing speed. Finally, we propose an even more compact neuron design which does not contain any pre-computer bank: a Multiplier-less Neuron, leading to large improvement in energy consumption with minimal accuracy degradation.

## II. ARTIFICIAL NEURAL NETWORK: BASICS

The fundamental elements of these artificial neural networks are neurons and synapses. When modelling artificial neurons, the complexity of a biological neuron is highly abstracted. The artificial neuron calculates a weighted sum of inputs and passes the result through an activation function. The activation function can be hard-limiting (e.g. step function) or soft-limiting (e.g. logistic sigmoid functions). Soft-limiting neurons (Fig.1 (a)) are preferred as they allow much more information to be communicated across neurons and greatly improve the neural network modeling capability while reducing network complexity. By adjusting the weights corresponding to the inputs of an artificial neuron, we can obtain desired output for specific inputs; this process is called training.

While our proposed approach can be applied to various classes of ANNs, in this work we consider the ubiquitous form, i.e. feedforward ANNs. In feedforward ANNs the neurons are connected in such a way that they form an acyclic network, this is illustrated in Fig.1 (b).

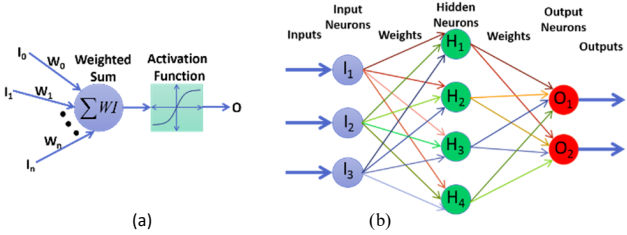

Fig. 1. (a) Artificial neuron, (b) Feedforward Artificial Neural Network. In (b), each neuron of one layer is connected to all the neurons of the following layer which is shown by arrows. The different colored arrows indicate that each input is multiplied by different weights.

The basic operation of these ANNs consists of two stages: i) Training and ii) Testing. The training process is usually carried out offline and therefore is not of much concern from energy consumption aspect. The trained ANN is then used to test random data inputs, and this is done on-chip. For large networks with millions of neurons, the testing process, although less compute-intensive than training, may also require significant computation. The testing process is basically forward propagation, which consists of multiplication, summation and activation operations. The most power consuming operation among these is the multiplication, which by far outweighs the summation and activation. And therefore, our main focus is to mitigate this issue by providing a solution which is energy efficient. In this work, we first replace the conventional multiplier in the neurons with approximate ASM. Finally we make an effort to create a multiplier-less artificial neuron. Note that, retraining the network with approximate multipliers leads to minimal degradation in accuracy while achieving significant energy reduction.

## III. ALPHABET SET MULTIPLIER

In a multiplication operation, the product can be generated from smaller bit sequences, which are the lower order multiples of the multiplier input '$I$'. The decomposition is based on the multiplicand '$W$', which in our case represents the synapse weights. Sample decompositions of two multiplication operations $W_1 \times I$ and $W_2 \times I$, are shown in Table I.

TABLE I. DECOMPOSITION OF MULTIPLICATION OPERATION

| Weights | Decomposition of Product |
|---|---|
| $W_1 = 01101001_2$ ($105_{10}$) | $W_1 \times I = 2^5.(0011).I + 2^0.(1001).I$ |
| $W_2 = 01000010_2$ ($66_{10}$) | $W_2 \times I = 2^6.(0001).I + 2^1.(0001).I$ |

Note that, if $I$, $3I$, $5I$, $7I$, $9I$, $11I$, $13I$, and $15I$ are available, the entire multiplication is reduced to a few shift and add operations. These smaller bit sequences, $a_k$ are referred to as alphabets. In ASM [30-32], instead of multiplying the multiplier with the multiplicand, some pre-specified alphabets are shifted and added. These alphabets are collectively called the alphabet set and consists of lower order multiples of the multiplier. A pre-computer bank is required to generate the alphabets. Overall, the ASM has four steps: i) generate the alphabets ii) select an alphabet iii) shift that alphabet iv) add the shifted alphabets. In this work, synapse weights are taken as 8(12) bit words and divided into two (three) quartets for the ASMs. This requires a final addition after select and shift operations. Based on the multiplicand, different combinations of select, shift and addition will occur which will be controlled by some control logic. In order to cover all possible combinations and to perform general multiplication operation, it has been shown that 8 alphabets {1,3,5,7,9,11,13,15} are required for bit sequence size of 4 bits [30]. It should be noted that the number of alphabets directly translates to power dissipation of the pre-computer unit, while the number of communication buses (out of the pre-computer) is also proportional to the number of alphabets. However, exploiting the error resilience of neural computing, the number of alphabets can possibly be reduced to achieve lower routing complexity and power dissipation. Using Fig. 2, the working principle of an 8 bit 4 alphabet ASM is explained next.

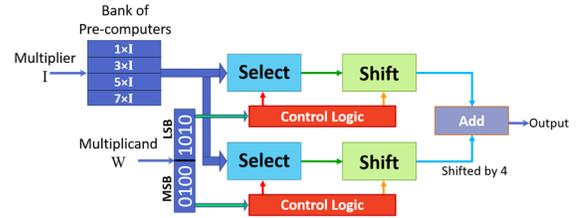

Fig. 2. 8 bit 4 alphabet ASM.

Multiplier '$I$' is fed to the pre-computer bank which generates 4 alphabets. In this example the alphabet set is {1,3,5,7}. Multiplicand '$W$' is divided into two parts which are the inputs of the 'control logic' circuits. Based on the '$W$', appropriate control logic for the '*select*' and '*shift*' units are generated. The select units select proper alphabets and pass them to the shift units. Shift units shift the input by the required amount. Finally, the '*adder*' unit adds the two separate values to get the multiplication result. For example, if the multiplier is '$M$' and multiplicand is $01001010_2$, we have to generate $1010_2\ M$ ($10M$) and $0100_2\ M$ ($4M$) $\times 2^4$ (shifted by 4 for MSB), and add them. $10M$ can be generated by shifting the alphabet $5M$ by 1. $4M$ can be generated by shifting the alphabet $1M$ by 2. The final addition is demonstrated by the following equation:

$$01001010_2 \times M = (4M) \times 2^4 + (10M) \times 2^0$$

These ASMs will only be advantageous if they can be used in a distributed way with minimum number of alphabets, i.e. share the alphabets with multiple multiplication units. The CSHM [33] architecture serves that purpose. Fig. 3 shows a CSHM consisting of a common pre-computer bank, shared between a number of multiplication units.

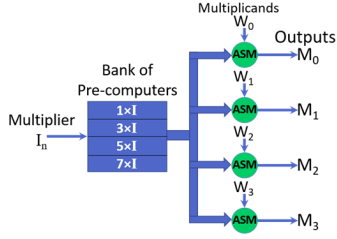

Fig. 3. 4 alphabet ASMs using CSHM architecture.

In a feedforward ANN, the pre-computer bank can be shared as each input is multiplied by a number of different weights to feed the different neurons (Fig. 1(b)). We implemented a processing unit that processes four neurons at a time, thus making it possible for 4 ASM units to share the alphabets from a common pre-computer bank, as illustrated in Fig. 3.

### IV. Design Approach & Methodology

The use of ASM to exploit error resilience, and sharing of alphabets are the bases of MAN. This section outlines the key ideas behind MAN and the proposed design methodology.

#### A. Introduction of Weight Constraints

The efficacy of ASM largely depends on the number of alphabets used to cover the range of combinations of select, shift and add. If the bit sequences used for the decomposition of the multiplication operation contains 4bits, then an alphabet set of 8 alphabets $\{1,3,5,7,9,11,13,15\}$ is sufficient to generate any product using the select, shift and add operations. To gain substantial performance improvements, we propose the use of reduced number of alphabets– in other words, we may not cover all the combinations, leading to approximations in multiplication. For example, if we use 4 alphabets $\{1,3,5,7\}$, we can generate 12 (including 0 ($0000_2$)) out of 16 possible combinations of 4bits by bit shift operations (e.g. from 1 ($0001_2$) we get 2 ($0010_2$), 4 ($0100_2$) and 8 ($1000_2$)). In this case, the unsupported bit quartet values are $\{9,11,13,15\}$. Therefore, we cannot generate the product $01101001_2 \times I$ with any select, shift and add combinations, as the LSB $1001_2(9_{10})$ is not supported by the used alphabet set. To alleviate this problem we introduce constrained training of the ANN so that these unsupported combinations never occur. Since ANN applications are error resilient, we can exploit this and get suitable set of weights while incurring minimum or no loss in network accuracy by retraining the network with the imposed constraints. The retraining overhead is minimal compared to the original training.

Next, the algorithm for constraining weights for 12 bit ASM is explained as an example. Consider the 12 bit synapse weight as a concatenated version of 3 bit quartets P, Q and R, where P is the MSB and R is the LSB as shown in Fig. 4. Since we are using 2's complement binary number system, the first bit of P is the sign bit; we do not have to consider that bit as we will multiply only the absolute value. So, P can have 8 combinations, 0 ($000_2$) to 7 ($111_2$), while Q and R can have 16 combinations, 0 ($0000_2$) to 15 ($1111_2$). If we use 2 alphabets $\{1,3\}$ only, the maximum number of supported combinations out of the 16 is 8. In that case, we cannot support 5 and 7 for P, while 5, 7, 9, 10, 11, 13, 14, 15 for Q and R. Hence, we convert those unsupported values to the nearest supported value ensuring minimum loss in precision. Algorithm 1 is the weight constraint mechanism for 12 bit 2 Alphabets $\{1,3\}$ Multiplier.

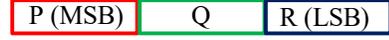

Fig. 4. 12 bit weight value decomposed into three quartets.

**Algorithm 1:** Weight constraint for 12 bit 2 Alphabets$\{1,3\}$ Multiplier
*Input:* Weight value *PQR*, list of unsupported quartets values $\{unsV\}_P$, $\{unsV\}_{Q,R}$
*Output:* Updated Weight value $PQR_{new}$
1. If $P=\{unsV\}_P$
2.    If $Q=\{unsV\}_{Q,R}$
3.       If $R=\{unsV\}_{Q,R}$, then round-down *R*,
4.             based on $R_{new}$ round-up/down *QR*,
5.             based on $Q_{new}$ round-up/down *PQR*
6.       Else based on *R* round-up/down *QR*,
7.             based on $Q_{new}$ round-up/down *PQR*
8.    Else based on *Q* round-up/down *PQR*
9. Else if $Q=\{unsV\}_{Q,R}$
10.   If $R=\{unsV\}_{Q,R}$, then round-down *R*,
11.            based on $R_{new}$ round-up/down *QR*,
12.             based on $Q_{new}$ round-up/down *PQR*
13.   Else based on *R* round-up/down *QR*,
14.            based on $Q_{new}$ round-up/down *PQR*
15. Else if $R=\{unsV\}_{Q,R}$, then round-down *R*,
16.            based on $R_{new}$ round-up/down *QR*,
17.            based on $Q_{new}$ round-up/down *PQR*
18. Return $PQR_{new}$

**Rounding Logic:** For correct multiplication operation, we must round-up/down an unsupported value to the nearest supported value ensuring minimum loss of information. For every two consecutive supported values, the average of them is considered as the threshold point for rounding. Consider the two consecutive supported values of 8 and 12 (using only the alphabets $\{1,3\}$); then the threshold is $(8+12)/2=10$. If the unsupported value 9 comes up, we will convert it to 8, else if 10 or 11 comes up, we will convert it to 12. The threshold point for rounding is different for different unsupported values.

#### B. Neural Network Design Methodology

With help of Fig.5, algorithm 2 describes the overall NN training and testing methodology. The inputs are a neural network (NN), its corresponding training dataset (TrData), testing dataset (TsData), and a quality constraint (Q) that dictates the degradation in quality tolerable in the implementation. The quality specifications are application-specific.

**Algorithm 2:** NN training and testing methodology
*Input:* Neural network: NN, Training dataset: TrData, Testing dataset: TsData, Quality constraint: $Q \leq 1$
*Output:* Retrained NN meeting the quality constraint.
1. Train the NN using TrData without any weight constraints till the training reaches near saturation, i.e. minuscule improvement in recognition accuracy can be achieved through more training.
2. Test the network using the TsData to get the network accuracy *J*. Create a restore point.
3. Retrain the network imposing constraints for minimum number of alphabets (start with 1) on weight update with lower learning rate till it again reaches near saturation.
4. Test the retrained network to find the new network accuracy *K* and compare the network accuracy using *J*, *K* and *Q*.
    If accuracy is satisfactory, i.e. if $K \geq J \times Q$, then end the training. Else restart from the restore point created in 2 and repeat steps 3 and 4 with increased number of alphabets.

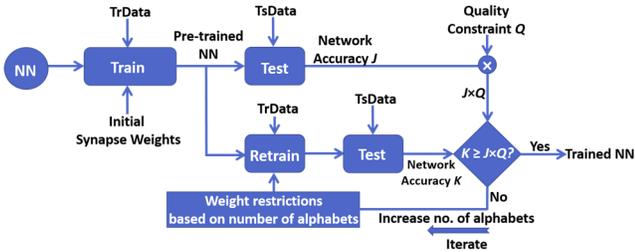

Fig. 5. Overview of the ANN design methodology.

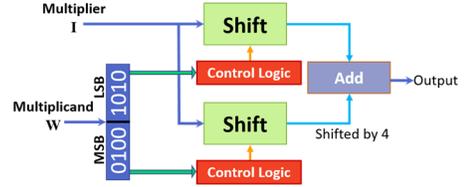

Fig. 6. 8 bit 1 alphabet {1} ASM.

## C. Credibility of the Proposed Design

To test the credibility of our design, we employed it on Face Detection application where based on input image data, the network detects whether there is a face present or not. Here, the number of final output neurons is only 2. We used 1024 input neurons and 100 hidden layer neurons. Using the training dataset, we first generated the 8 and 12 bit synapse weights for unconstrained (for conventional multiplier) and constrained conditions (for ASM). Then we tested the network using the test dataset and achieved good results with a maximum degradation in accuracy of 0.47%. The results are listed in Table II.

TABLE II.   NN ACCURACY RESULTS FOR FACE DETECTION

| Size of Synapse | No. of Alphabets | Accuracy (%) | Accuracy Loss (%) |
|---|---|---|---|
| 8 bits | conventional NN | 90.66 | -- |
|  | 4 {1,3,5,7} | 90.46 | 0.22 |
|  | 2 {1,3} | 90.31 | 0.39 |
|  | 1 {1} | 90.23 | 0.47 |
| 12 bits | conventional NN | 90.71 | -- |
|  | 4 {1,3,5,7} | 90.60 | 0.12 |
|  | 2 {1,3} | 90.54 | 0.19 |
|  | 1 {1} | 90.49 | 0.24 |

*Accuracy loss is computed by considering the conventional NN accuracy as standard.

After this success, we moved on to a more complex problem of 'Hand Written Digit Recognition' using MNIST [34] dataset. We used similar method as before to generate the synapse weights (here, the number of final output neurons is 10). Then we used those synapse weights in our designed processing engine to test the network accuracy. The accuracy results are listed in Table III.

TABLE III.   NN ACCURACY RESULTS FOR DIGIT RECOGNITION

| Size of Synapse | No. of Alphabets | Accuracy (%) | Accuracy Loss (%) |
|---|---|---|---|
| 8 bits | conventional NN | 97.45 | -- |
|  | 4 {1,3,5,7} | 97.41 | 0.04 |
|  | 2 {1,3} | 97.39 | 0.06 |
|  | 1 {1} | 97.11 | 0.35 |
| 12 bits | conventional NN | 97.63 | -- |
|  | 4 {1,3,5,7} | 97.60 | 0.03 |
|  | 2 {1,3} | 97.44 | 0.19 |
|  | 1 {1} | 97.38 | 0.25 |

## D. Multiplier-less Neuron

Getting the results using ASM in artificial neurons, we observed that even with only 1 alphabet {1} in all layers, we are able to achieve network accuracy within ~0.5% of conventional implementation. The added advantage of using only 1 alphabet, specifically {1}, is that we do not have to generate and use any alphabet set, the input only is sufficient for the 1 {1} alphabet requirement. That means we do not need multiplication, only shifting and adding is enough. This eliminates the necessity of the pre-computer bank and alphabet 'select' unit (Fig. 6). Hence, the circuit would be faster, more compact and less power consuming, leading to a 'Multiplier-less' neuron.

## V. SIMULATION FRAMEWORK

This section describes the overall simulation framework. We used multilayer perceptron models and convolutional neural networks for our experiment. We used modified versions of the open source C++ [35] and MATLAB [36] codes. Using these we implemented multi-layer backpropagation networks. We trained the NNs using the corresponding training datasets. Then restrictions in the weight update were imposed during retraining of the NNs, so that the reduced number of alphabets in ASM based neurons can be used. These synapse weights from the trained NNs along with the test patterns were used as inputs for our processing engine. The processing engine was implemented at the Register-Transfer Level (RTL) in Verilog and mapped to the IBM 45nm technology using Synopsys Design Compiler Ultra. It was also used to estimate the energy consumption and area under iso-speed conditions. The metrics for the benchmarks used are listed in Table IV:

TABLE IV.   BENCHMARKS

| Application | Dataset | NN Model | No. of Layers | No. of Neurons | No. of Trainable Synapses |
|---|---|---|---|---|---|
| Digit Recognition(8bit) | MNIST | MLP | 2 | 110 | 103510 |
| Digit Recognition(12bit) | MNIST | CNN (LeNet) [37] | 6 | 8010 | 51946 |
| Face Detection (12bit) | YUV Faces | MLP | 2 | 102 | 102702 |
| House Number Recognition | SVHN | MLP | 6 | 1560 | 1054260 |
| Tilburg Character Set Recog. | TICH | MLP | 5 | 786 | 421186 |

The key implementation metrics are shown in Table V.

TABLE V.   EXPERIMENTAL PARAMETERS

| Metric | Value |
|---|---|
| Feature Size | 45nm |
| Clock Frequency for 8 bits Neuron | 3 GHz |
| Clock Frequency for 12 bits Neuron | 2.5 GHz |

## VI. RESULTS

In this section, we present results that demonstrate the accuracy obtained and also the energy efficiency and area reduction achieved by our proposed design.

## A. Accuracy Comparison

Fig. 7 shows the classification accuracy obtained using conventional multiplier based neurons and proposed ASM based neurons for various applications. The accuracy of the proposed design in each application is normalized to a fully-accurate NN implementation in which all of the neurons use conventional multiplier. Note that the baseline for the networks are highly optimized to begin with since we are using only 8-12 bits for input and synapse values. The maximum loss in accuracy was ~2.83% and ~0.25% for 8 bit and 12 bit neurons respectively.

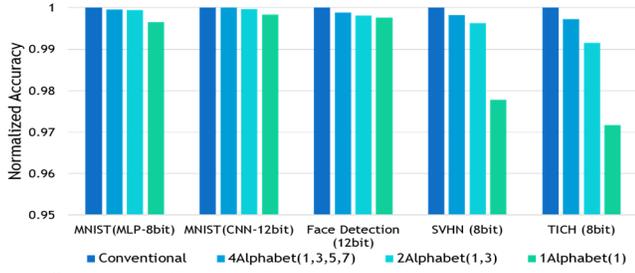

Fig. 7. Comparison of accuracy between conventional multiplier based NN and ASM based NNs.

The classification accuracy of ASM based NNs is very good for simple datasets such as MNIST and YUV Faces, compared to more complex datasets such as SVHN and TICH.

### B. Power Benefits and Energy Consumption Comparison

Fig. 8 shows the average power improvement achieved using ASMs for different size (bit-widths) of neurons. The power consumption of each scheme is normalized to an implementation in which all the neurons utilize conventional multipliers.

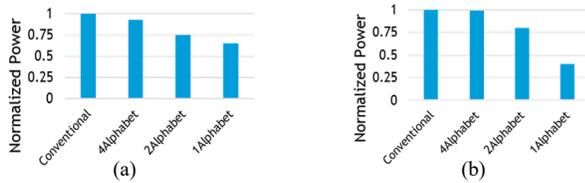

Fig. 8. Comparison of power consumption, between conventional neuron and different ASM based neurons of (a) 8 bit and (b) 12 bit.

For 8 bit neurons we achieve ~8-26% reduction in power consumption using 4 {1,3,5,7} and 2 {1,3} alphabets, respectively. For 12-bit neurons, we get up to ~21% reduction in power consumption using only 2 {1,3} alphabets. In the case of multiplier-less neurons, we achieve ~35% and ~60% reduction in power consumption, respectively, for 8 bit and 12 bit neurons, using only 1 alphabet {1}. Fig. 9 shows the energy savings obtained using ASMs for different applications grouped by the size and type of NN. The amount of energy savings increases almost linearly with the increase in NN size.

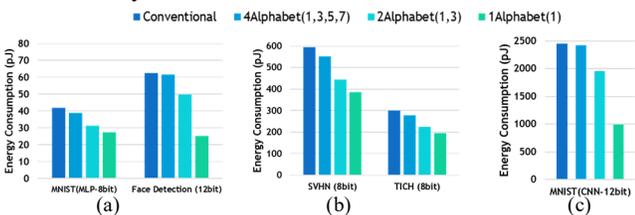

Fig. 9. Comparison of energy consumption, between conventional neuron and different ASM based neurons in (a) 2 layer MLPs, (b) 5-6 layer MLPs and (c) 6 layer CNN.

### C. Area Reduction

Fig. 10 shows the area reduction obtained using ASMs. Again, the area of each scheme is normalized to a conventional neuron implementation in which all the neurons utilize conventional multipliers. For a neuron size of 8 bits we achieve ~5-25% reduction in area using 4 {1,3,5,7} and 2 {1,3} alphabets, respectively. For 12-bit neurons, we get up to ~19% reduction in area using only 2 {1,3} alphabets. In the case of multiplier-less neurons, we achieve ~37% and ~62% reduction in area, respectively, for 8 and 12 bit neurons, using only 1 alphabet {1}.

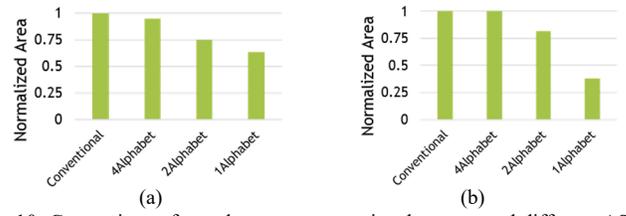

Fig. 10. Comparison of area between conventional neuron and different ASM based neurons of (a) 8 bit and (b) 12 bit.

### D. Overall Improvements

The results show that, for 4 {1,3,5,7} alphabet ASM based neurons, we may not achieve significant improvement in terms of power and area. But using 2 {1,3} alphabet ASM based neurons, we can get up to ~26% reduction in power consumption and ~25% reduction in area with a ~0.85% loss in accuracy. Whereas in the case of MAN, we achieve ~35% and ~60% reduction in power consumption, and ~37% and ~62% reduction in area, respectively, for 8 and 12 bit neurons, using only 1 alphabet {1}, with a maximum ~2.83% loss in accuracy. These comparisons were performed under iso-speed conditions with clock frequency of 3 GHz and 2.5 GHz for 8 and 12 bit neurons, respectively.

### E. Add-on Accuracy Improvement Through Retraining

From Fig.7 it can be observed that 12 bit neurons provide much better network accuracy even with only one alphabet{1}, with a maximum ~0.25% degradation in accuracy. As 12 bit synapses has more flexibility compared to 8 bit synapses, the NN can be retrained better to compensate for the reduced number of alphabets in ASM based neurons.

On the other hand 8 bit synapses demonstrate considerable degradation when only one alphabet{1} is used, with a maximum ~2.83% loss in accuracy. This shortcoming can be tackled by using more number of alphabets in small number of significant neurons. Usually NNs have smaller numbers of neurons in the concluding layers. Also, it has been shown that these neurons have more influence in determining the final output of the NN compared to the neurons in initial layers [29]. Exploiting this insight, we can use one alphabet{1} in the initial larger layers, and two alphabets{1,3} or four alphabets{1,3,5,7} in the ending smaller layers to improve the network accuracy. This will also increase the energy consumption as two alphabets{1,3} and four alphabets{1,3,5,7} based ASMs consume much more power than one alphabet{1} based ASM. But this increase is quite small in practice as the ending smaller layers with fewer neurons account for a very small percentage of total processing cycles of the NN. For example, in the 6 layer 'House number recognition' network, the last 2 layers use only 3.84% of total processing cycles. In Fig.11, the efficacy of this technique is illustrated. For handwriting recognition of the MNIST dataset using a 2 layer MLP network, 1 alphabet ASM based neurons are used in the only hidden layer and 4 alphabet ASM based neurons are used only in the output layer. For recognition of the SVHN dataset using a 6 layer network, 1 alphabet ASM based neurons are used in the first four hidden layers, and 2 and 4 alphabet ASM based neurons are used in the penultimate and ultimate layer, respectively. For recognition of the TiCH dataset using a 5 layer network, 1 alphabet ASM based neurons are used in the first three hidden layers, and 2 and 4 alphabet ASM based neurons are used in the penultimate and ultimate layer, respectively.

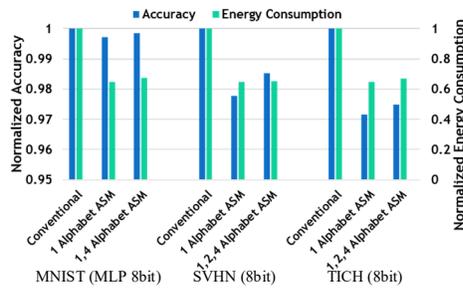

Fig. 11. Comparison of accuracy and energy consumption between conventional NN, 1 alphabet ASM based NN and 1,2,4 alphabet ASM based NN. For 1 alphabet ASM, 2 alphabets ASM and 4 alphabets ASM, the alphabets used are {1}, {1,3} and {1,3,5,7} respectively.

From Fig.11, we can observe that, in all the three applications, the accuracy is improved using higher number of alphabets in the neurons in the concluding layers with very small energy overhead. Therefore, for NNs with neuron size constraints, this method can be employed to obtain a better tradeoff between energy and accuracy.

## VII. Conclusion

Large-scale neural networks have attracted great interest in a wide range of applications. However, the ever-growing complexity of networks and dataset sizes place significant demands on computing platforms. In this work, we exploited the resilience of neural network applications to design highly efficient and approximate ASM based neurons, in order to achieve energy benefits. We further proposed the concept of Multiplier-less Artificial Neuron (MAN), in which the conventional multiplier is replaced by the most simplified shift and add operations. We retrained the approximate networks with the weight constraints, providing the opportunity to mitigate the accuracy loss due to neuron approximation. Our experiments demonstrated significant improvements in energy consumption and reduction in area for negligible loss in the classification accuracy.

## Acknowledgment

This work was supported in part by C-SPIN, one of six centers of STARnet, a Semiconductor Research Corporation program, sponsored by MARCO and DARPA, by the Semiconductor Research Corporation, by the National Science Foundation, and by Intel Corporation.

## References

[1] Y.Özbay & G. Tezel, "A New Method for Classification of ECG Arrhythmias Using Neural Network with Adaptive Activation Function," *Digital Signal Processing* 20.4, pp.1040-1049, 2010.

[2] G. Dede & M. H. Sazlı, "Speech Recognition with Artificial Neural Networks," *Digital Signal Processing* 20.3, pp.763-768, 2010.

[3] A. Marcano-Cedeño et al. "WBCD Breast Cancer Database Classification Applying Artificial Metaplasticity Neural Network," *Expert Systems with Applications* 38.8, pp. 9573-9579, 2011.

[4] A. Krizhevsky et al. "ImageNet Classification with Deep Convolutional Neural Networks," *In Advances in neural information processing systems*, pp. 1097-1105, 2012.

[5] K. He et al. "Spatial Pyramid Pooling in Deep Convolutional Networks for Visual Recognition," In *Computer Vision–ECCV*, pp. 346-361, Springer International Publishing, 2014.

[6] Y. Sun et al. "DeepID3: Face Recognition with Very Deep Neural Networks," *arXiv preprint arXiv:1502.00873*, 2015.

[7] T. Chilimbi et. al. "Project Adam: Building an Efficient and Scalable Deep Learning Training System," In *11th USENIX Symposium on OSDI*, pp. 571-582, Oct. 2014.

[8] F. M. Dias et al. "Artificial Neural Networks: A Review of Commercial Hardware," *Engineering Applications of Artificial Intelligence* 17.8, pp. 945-952, 2004.

[9] J. Misra et al. "Artificial Neural Networks in Hardware: A Survey of Two Decades of Progress," *Neurocomputing*, 74.1, pp. 239-255, 2010.

[10] C. Farabet et al. "Neuflow: A Runtime Reconfigurable Dataflow Processor for Vision," In *Proc. CVPRW*, pp. 109-116, IEEE, June 2011.

[11] N. Sudha et al. "A Self-Configurable Systolic Architecture for Face Recognition System Based on Principal Component Neural Network," *IEEE Trans. on Circuits and Systems for Video Technology*, 21.8, pp. 1071-1084, 2011.

[12] S. Chakradhar et al. "A Dynamically Configurable Coprocessor for Convolutional Neural Networks," In *ACM SIGARCH Computer Architecture News*, 38.3, pp. 247-257, ACM. June 2010.

[13] J. Ngiam et al. "On Optimization Methods for Deep Learning," In *Proc. ICML*, pp. 265-272, 2011.

[14] R. V. Hoang et al. "A Novel CPU/GPU Simulation Environment for Large-Scale Biologically Realistic Neural Modeling," *Frontiers in neuroinformatics* 7, 2013.

[15] K. H. Kim et al. "A Functional Hybrid Memristor Crossbar-Array/CMOS System for Data Storage and Neuromorphic Applications," *Nano letters*, 12.1, pp. 389-395, 2011.

[16] M. Suri et al. "Phase Change Memory as Synapse for Ultra-Dense Neuromorphic Systems: Application to Complex Visual Pattern Extraction," In Proc. *IEDM*, *IEEE*, pp. 4-4. IEEE, Dec. 2011.

[17] C. D. Wright et al. "Beyond Von-Neumann Computing with Nanoscale Phase-Change Memory Devices," *Advanced Functional Materials*, 23.18, pp. 2248-2254, 2013.

[18] B. Rajendran et al. "Specifications of Nanoscale Devices and Circuits for Neuromorphic Computational Systems," *IEEE Trans. on Electron Devices*, 60.1, pp. 246-253, 2013.

[19] A. Sengupta et al. "Spin-Orbit Torque Induced Spike-Timing Dependent Plasticity," *Applied Physics Letters*, 106.9, 093704, 2015.

[20] K. Roy et al. "Beyond Charge-Based Computation: Boolean and Non-Boolean Computing with Spin Torque Devices," In *Proc. ISLPED*, pp. 139-142, IEEE, Sept. 2013.

[21] M. Sharad et al. "Proposal For Neuromorphic Hardware Using Spin Devices," *arXiv preprint arXiv:1206.3227*, 2012.

[22] V. K. Chippa et al. "Scalable Effort Hardware Design: Exploiting Algorithmic Resilience for Energy Efficiency," In *Proc. DAC*, pp. 555-560, ACM, 2010.

[23] S. Venkataramani et al. "SALSA: Systematic Logic Synthesis of Approximate Circuits." In *Proc. DAC*, pp. 796-801, ACM, June 2012.

[24] V. Gupta et al. "IMPACT: Imprecise Adders for Low-Power Approximate Computing," *In Proc. ISLPED*, pp. 409-414, IEEE Press, Aug. 2011.

[25] H. Cho et al. "ERSA: Error Resilient System Architecture for Probabilistic Applications," *IEEE Trans. on Computer-Aided Design of Integrated Circuits and Systems*, 31.4, pp. 546-558, 2012.

[26] Y. Kim et al. "An Energy Efficient Approximate Adder with Carry Skip for Error Resilient Neuromorphic VLSI Systems," In *Proceedings of the International Conference on Computer-Aided Design*, pp. 130-137, 2013.

[27] H. Esmaeilzadeh et al. "Neural Acceleration for General-Purpose Approximate Programs," In *Proceedings of the 2012 45th Annual IEEE/ACM International Symposium on Microarchitecture*, pp. 449-460, *IEEE Computer Society*, 2012.

[28] Y. Yetim et al. "Extracting Useful Computation from Error-Prone Processors for Streaming Applications," In *Proc. DATE*, pp. 202-207, IEEE. March 2013.

[29] S. Venkataramani et al. "AxNN: Energy-Efficient Neuromorphic Systems Using Approximate Computing," In *Proc. ISLPED,* pp. 27-32, ACM, 2014.

[30] S. Sivanantham et al., "Low Power Floating Point Computation Sharing Multiplier for Signal Processing Applications," *International Journal of Engineering and Technology (IJET)* 5.2, pp. 979-985, 2013.

[31] G. Karakonstantis & K. Roy, "An Optimal Algorithm for Low Power Multiplierless FIR Filter Design Using Chebychev Criterion," In *Proc. ICASSP*, Vol. 2, pp. II-49, IEEE, April 2007.

[32] J. Park, et al. "Non-Adaptive and Adaptive Filter Implementation Based on Sharing Multiplication," In *Proc. ICASSP*, Vol. 1, pp. 460-463, IEEE, 2000.

[33] K. Muhammad, "Algorithmic and Architectural Techniques for Low Power Digital Signal Processing," PhD thesis, Purdue University, 1999.

[34] http://yann.lecun.com/exdb/mnist/

[35] https://github.com/nyanp/tiny-cnn

[36] https://github.com/rasmusbergpalm/DeepLearnToolbox

[37] Y. LeCun et al. "Gradient-based learning applied to document recognition," *Proceedings of the IEEE*, 86.11, 2278-2324, 1998